\documentclass[11pt
]{ceurart}

\newcommand\nnfootnote[1]{%
  \begin{NoHyper}
  \renewcommand\thefootnote{}\footnote{#1}%
  \addtocounter{footnote}{-1}%
  \end{NoHyper}
}

\begin{document}

\copyrightyear{2023}
\copyrightclause{Copyright for this paper by its authors.
  Use permitted under Creative Commons License Attribution 4.0
  International (CC BY 4.0).}

\conference{AutoMate'23@IJCAI: The First International Workshop on the Future of No-Code Digital Apprentices, Aug 21st, 2023, Macao S.A.R.}

\title {Enhancing Trust in LLM-Based AI Automation Agents: New Considerations and Future Challenges}

\author{Sivan Schwartz*}[
email=Sivan.Schwartz@ibm.com,
]
\author{Avi Yaeli*}[
email=aviy@il.ibm.com,
]
\author{Segev Shlomov}[
email=Segev.Shlomov1@ibm.com,
]
\address{IBM Research}

\nnfootnote{* These two authors contributed equally to this work}

\begin{abstract}
Trust in AI agents has been extensively studied in the literature,
resulting in significant advancements in our understanding of this field. However, the rapid advancements in Large Language Models (LLMs) and the emergence of LLM-based AI agent frameworks pose new challenges and opportunities for further research. In the field of process automation, a new generation of AI-based agents has emerged, enabling the execution of complex tasks. At the same time, the process of building automation has become more accessible to business users via user-friendly no-code tools and training mechanisms. This paper explores these new challenges and opportunities, analyzes the main aspects of trust in AI agents discussed in existing literature, and identifies specific considerations and challenges relevant to this new generation of automation agents. We also evaluate how nascent products in this category address these considerations. Finally, we highlight several challenges that the research community should address in this evolving landscape.
\end{abstract}


\maketitle

\section{Introduction}
Trust, according to the Cambridge dictionary definition, is “To believe that someone is good and honest and will not harm you, or that something is safe and reliable”. Over the years, researchers, mainly in psychology and social economy disciplines, have researched and focused on the definition of trust and what attributes influence the magnitude and occurrence of trust between humans \cite{simpson2007psychological}. 
Trust in AI agents has also been extensively studied in research, resulting in significant advancements in our understanding of this field. Yang and Wibowo\cite{yang2022user} review current studies, components, and frameworks, for example, the trust framework for IT artifacts, as adopted by \cite{schmidt2020transparency}, which focuses on performance, process, and purpose categories as key dimensions for achieving trust in AI technology.  Each category is then further divided into specific trust components such as degree of responsibility with accurate information, reliability considering the user's perception of consistency in completing a given task, predictability, and faith.

In the last couple of years, the field of artificial intelligence (AI) has been advancing at an exponential rate.  Large language models (LLMs) based on generative pre-trained transformer architectures such as OpenAI’s GPT-3 \cite{dale2021gpt}, and more recently ChatGPT \cite{ChatGPT}'s and GPT-4  \cite{adesso2022gpt4} have shown astonishing capabilities in understanding and generating language, meeting or exceeding expert-level skills in content generation, question answering, and even discovering new scientific knowledge.  More recently, advanced frameworks have emerged that enable the creation of autonomous AI agents.  AutoGPT \cite{AutoGPT}, AgentGPT \cite{AgentGPT}, LangChain \cite{LangChain}, and ChatGPT Plugins \cite{ChatGPT-Plugins}, are capable of receiving high-level objectives, decomposing them into sub-tasks, and interacting online with applications, services, APIs, and tools.  

These new capabilities set the stage for a new generation of AI-based agents that can be autonomous, generate thoughts and observations, make decisions, interact with the world, and perform practically any type of complex task without the need for humans to create prompts.  Products such as Microsoft CoPilot \cite{MicrosoftCopilot}, Adept.AI \cite{AdeptAct1}, MultiOn \cite{MultiOn}, chatGPT, and Watson Orchestrate \cite{WatsonOrchestrate} have begun to embed these models and frameworks while providing AI-assistants that help humans perform work.   These examples mark an era of highly advanced digital assistants that are integrated into the tools people use and provide superpower skills for data entry, analysis, process automation, content generation, and more.  In most of these examples, the human is still in the loop (human-in-the-loop) in order to control and supervise the AI model.

However, in the context of business process automation, AI-based agents will also be poised to make automatic decisions and take actions, often on increasingly more complex types of tasks.  At the same time, the process of building these automations is becoming more accessible to business users via user-friendly no-code tools and training mechanisms.  In these settings, how can we ensure that AI-based agents will not make harmful mistakes, compromising human and enterprise trust that can not be easily recovered?  

Trust is a prerequisite to technology adoption. In the early days of the Internet, people did not agree to conduct financial transactions due to the lack of many components of trust.  The IT industry had to develop mechanisms and technologies to build trust, such as secure transmission protocols, encryption algorithms to protect user-sensitive data and privacy, and security threat mitigation tools. We should assume that these ways of building trust with users will also be important for the adoption of AI-based Automation agents.  In addition, since LLMs are now capable of comprehending and generating language much like humans, additional considerations of trust that have not been the focus of previous frameworks may require attention.  We therefore ask, in this quickly evolving landscape of AI-based automation agents, what new challenges and opportunities lie ahead for research?

This paper explores these new challenges and opportunities.  In Section \ref{sec:trust} we present key categories of trust, and highlight specific considerations and challenges for LLM-based AI automation agents.  In Section \ref{sec:assessment}, we evaluate several nascent products and assess the extent to which they consider the topics described above.  Finally, in Section  \ref{sec:discussion}, we summarize key challenges that the research community should address.

\section{Trust in AI-based Automation Agents}
\label{sec:trust}

Researchers and theoreticians from the psychology discipline have devoted a great effort to the concept of trust in human-to-human interaction, which serves as a crucial factor for fruitful well-functioning relationships and development \cite{simpson2007psychological}. From a dyadic (interpersonal) perspective \cite{holmesjk}, trust can be understood as a psychological state or inclination of an individual (referred to as the truster) towards a particular partner (known as the trustee) with whom they share interdependence. In this context, interdependence refers to a situation where the truster relies on the trustee's cooperation to achieve desired outcomes. Several major theories in psychology, such as Erick Erickson's theory of development \cite{erickson1963childhood} and John Bowlby’s theory of attachment \cite{bowlby1969disruption}, also highlight the importance of high levels of trust early in life as it predicts future better human-to-human functioning relationships. 

Since trust plays a crucial role in human interaction, researchers in the field also focus on exploring what factors affect trust levels. 
Trust can be divided into two categories, (1) cognitive trust, and (2) emotional trust. (1) Cognitive trust refers to a trust that is based on rational judgments and evaluations of another person's reliability, competence, and integrity. It involves the cognitive processes of gathering and analyzing information to make reasoned judgments about the trustworthiness of others. (2) Emotional trust, on the other hand, is based on affective or emotional bonds between individuals. It involves forming an emotional connection, empathy, and the perception of shared values, beliefs, and goals \cite{cook1980new}\cite{johnson2005cognitive}. In this paper, we will address both aspects of trust synergistically. 

As mentioned before, an excessive amount of research in psychology has focused on the factors which affect the trust level. Some of the main factors that are shown empirically to affect interpersonal trust levels are \emph{predictability} (i.e., consistent behavior of the trustee that enables the truster to predict the outcome of his actions), \emph{dependability} (i.e., reliability and accuracy of the trustee), and \emph{faith} (i.e., commitment to shared goals and values such that the truster will give a leap of faith to the trustee) \cite{rempel1985trust}\cite{johnson1982measurement}. These three factors are wide concepts that can be further divided into sub-factors such as consistency, transparency, commitment, etc. 

These factors affecting trust (both cognitive and emotional) in human-to-human interaction can be generalized and adapted to human-to-AI-agent interaction. After all, norms are shown to be shared between human-to-human and human-to-virtual/non-human agents \cite{madhavan2007similarities}. Moreover, as we reach the point where AI agents have the ability to understand and produce human natural language, key factors of human-to-human interaction for influencing trust are becoming even more relevant. Several studies have been conducted over the years in this field of human-to-virtual/non-human entities, highlighting and experimentally testing the factors which are important to build trust, originating from human-to-human trust studies. The main researched factors are described below.

\subsection{Reliability}\label{Reliability}
In the realm of virtual AI, users' trust and trusting behaviors are significantly influenced by the aspect of reliability. Empirical evidence supporting this notion can be found in the work of Moran and colleagues \cite{moran2013team}, who investigated compliance with voiced agent instructions in a team-based game. The researchers observed a strong correlation between trust in the agent and compliance with the instructions. When agent reliability was compromised, trust levels declined, reducing compliance. This finding underscores the critical role that reliability plays in fostering trust and ensuring successful interactions between users and AI. In addition, transparency regarding AI’s possible errors (leading to lower reliability expectations from the user) may increase trust in such scenarios. Moreover, in human-to-human interaction, consistent but also accurate responses are correlated with higher levels of trust \cite{jiang2014social}. The same pattern can be found in human to AI agent interactions \cite{dzindolet2002perceived}.



In the context of automation agents, reliability plays a crucial role in performing automated tasks.  To establish trust, it is imperative that agents consistently deliver high-quality results with a high degree of accuracy.  However, LLM-based agent frameworks could easily produce unanticipated outcomes stemming from either the infinite number of possible inputs by end users or the specific training and configuration done by non-technical users. How can we make sure that agents do not make harmful mistakes that would make human and enterprise trust difficult to recover? To achieve reliability the following dimensions require special consideration and attention:  

\begin{itemize}
  \item \textbf{[R-POM]: Prompt Mediation}.  Prompt mediation ensures that the user prompt does not reach the LLM before some level of quality and validation is performed. 
  \item \textbf{[R-COM]: Content and Output Mediation}. Content mediation guards the output from the LLM before it reaches the user.  This mechanism can assist in preventing unexpected behavior and meeting user expectations, such as preventing harmful actions or unwanted content.
  \item \textbf{[R-TG]: Task Grounding}.  Task grounding refers to the ability of the agent to recognize that the user intent belongs to a known and approved task to perform. This applies both to single tasks as well as the composition of tasks. 
  \item \textbf{[R-KG]: Knowledge Grounding} - knowledge grounding refers to the ability of the agent to recognize and validate that the entities mentioned by the user relate to specific parameters of the task and specific artifacts in the workspace of the user.   
  \item \textbf{[R-AG]: Application Grounding}.  Application grounding means that the agent understands the applications, services, and tools mentioned by the user, and can anticipate the outcome of using these applications, including downstream reliability.
  \item \textbf{[R-UF]: User Feedback}.  User feedback relates to the ability of the agent to take user responses into account. Once the user has provided corrective feedback, the agent is able to remember this for future interaction.  
  \item \textbf{[R-T]: Testing}.  Testing relates to (1) new testing techniques to address reliability in automation agents, and (2) the prevalence of a "dev" (development) mode for end users to validate the agent behavior against simulated business situations, prior to use in a real business setting.   
\end{itemize}

\subsection{Openness}\label{openess}
Openness, which includes transparency regarding goals, capabilities, and algorithmic workings, holds special importance for LLM-based AI automation agents. By clearly communicating the goals, users can have a better understanding of what the AI agent can and cannot do, thereby managing their expectations effectively and supporting the building of their trust.  In the context of being transparent about the agent's reliability, Fan and colleagues \cite{fan2008influence} found that informing participants of the actual reliability of a decision-making agent enhanced participants' trust and their performance in the task presented. 
Even in cases where the transparent reliability was low, participants more adeptly tailored their decisions, utilizing the agent's advice only when fitting. In other words, transparency plays a more critical role for trust than reliability in this context.

In the context of LLM-based automation agents, we identify the following dimensions to consider:

\begin{itemize}
    \item \textbf{[O-R]: Skills}. Disclosing to the user what skills the agent possesses and the level of reliability, consistency, and accuracy there is globally on those specific skills. 
    \item \textbf{[O-G]: Goals}.  Sharing the goal(s) of the agent and of the vendor that built the agent. 
    \item \textbf{[O-ALG]: Algorithms}.  Disclosing in non-technical terms how the algorithms work, and what data was used to train the models. 
    \item \textbf{[O-EL]: Ethics}.  Presenting any ethical or legal implications if they exist.
\end{itemize}

\subsection{Tangibility}\label{Tangibility}
In biology, phenotype matching refers to an implicit evaluation of relatedness based on some trait assessment of phenotypic similarity \cite{waldman1988ecology}. This similarity assessment may be either with reference to one’s own phenotypes or with those of one’s wider close population. This process has been found useful in the understanding of many species \cite{alberts1999paternal}\cite{mateo2000kin}. In humans, it has been shown that similarity, driven by phenotype matching behavior, enhances trust such that the more similar a person's facial appearance is to someone else's, the trust between the individuals increases \cite{debruine2002facial}.  In the context of AI agents, tangibility relates to the human user's tendency to trust agents with some aspects of similarities to themselves. As such, human-like agent characteristics are shown to increase participants’ apriori trust in the agent \cite{chattaraman2014virtual}. 

While LLM agents primarily interact through text or voice, incorporating tangibility through avatars or visual representations enhances the user experience by providing a more relatable and engaging interface. A human-like avatar appearance can facilitate better communication, as users tend to respond more positively to visually recognizable entities. It creates a sense of familiarity and relatability, fostering a stronger connection between the user and the AI agent. Tangibility also helps in conveying non-verbal cues and emotions, enriching the interaction and making it more intuitive. By incorporating avatar-like appearances, LLM-based AI automation agents can create a more immersive and satisfying user experience, improving communication and building trust with users.  To achieve tangibility, the following dimensions require special consideration and attention:

\begin{itemize}
    \item \textbf{[T-VR]: Visual Representation}.  The agent has a visual (graphical) representation, e.g. in the form of an avatar.
    \item \textbf{[T-NVC]: Non Verbal Cues}.  Non-verbal cues can include elements such as communicating the state of interaction channels (idle, listening, working, speaking) or communicating general emotions.  
\end{itemize}

\subsection{Immediacy behaviors}\label{Immediacy behaviors}
Immediacy behaviors refer to the perceived physical and/or psychological closeness between people \cite{mehrabian1967attitudes}. Immediacy behaviors can be defined as verbal and nonverbal communicative actions that send positive messages of liking and closeness, decrease the psychological distance between people, and positively affect motivation, according to past research. Such behaviors in the context of AI agents can increase their perceived anthropomorphism. Anthropomorphism refers to the innate tendency of humans to attribute human-like qualities or behavior to entities that are not human, such as animals, objects, or natural phenomena. It involves ascribing human characteristics or emotions to these non-human entities.

Previous research has indicated that instructor immediacy is associated with learning outcomes, satisfaction, motivation, and engagement \cite{schutt2009effects}. Social responsiveness and personalization of the virtual AI agent’s reactions can serve as immediacy behaviors for the human-to-AI agent interaction. Moreover, Pro-social virtual AI agent behaviors can be translated to perceptions of the agent’s personality \cite{andrews2012system}. Adding features in the AI agent to increase its anthropomorphic qualities may set unrealistic human expectations from the agent and therefore lead to a trust failure \cite{culley2013trust}\cite{mimoun2012case}. In the context of LLM-based automation, we highlight the following main considerations:  

\begin{itemize}
    \item \textbf{[I-EM]: Empathy}.  Using conversational language to express empathy, for example saying "I'm sorry, .." when making a mistake or not meeting the user's expectation.  This behavior can create a more human-like experience. 
    \item \textbf{[I-SA]: Style Adaptation}.  Adapting to the user's communication and working style can create a more personal and effective experience, ultimately improving user satisfaction and building stronger relationships with users.
\end{itemize}

\subsection{Task Characteristics}\label{Task Characteristics}

The task characteristics of LLM-based AI automation agents, including their ability to read or write data and the complexity of the task, hold special importance. When it comes to connecting to other systems, reading data is generally considered safer as it involves retrieving information without making any modifications.  On the other hand, writing data requires caution and often necessitates human-in-the-loop validation to ensure accuracy and avoid unintended consequences.  Additionally, more complex tasks often involve multiple agents or skills, each with its own trust characterization. In such cases, aggregating the outputs of these multiple agents becomes crucial, as it affects the overall reliability and performance of the system. As the sensitivity of the task increases, earning trust becomes more challenging, and maintaining it becomes more difficult \cite{glikson2020human}.  On the other hand, in some cases such as simple tasks that provide technical information, responses from agents were shown to be more trustworthy than those from humans \cite{glikson2020human}.  By carefully considering the task characteristics, LLM-based AI automation agents can adjust their responses to different types of tasks.  We consider the following main characteristics:

\begin{itemize}
    \item \textbf{[TC-HLM]: Human-in-the-loop Moderation}.  Agents that produce output or action that is guarded by a human reading the output or approving the action.
    \item \textbf{[TC-AA]: Autonomous Actions}. Agents that perform autonomous actions that may influence the business without any human supervision.
    \item \textbf{[TC-OET]: Open-ended Tasks}.  Agents that perform open-ended tasks, with no predefined condition or criteria to determine whether the task is properly completed.
\end{itemize}
 trust failure \cite{culley2013trust} \cite{mimoun2012case}  (Culley and Madhavan, 2013, Ben Mimoun et al., 2012).
 
\subsection*{Trust trajectory}
Although not a dimension for increasing trust levels, the trajectory of trust development is an intrinsic part that is shared for all dimensions influencing trust. 
Trust in the interaction of human to human or human to AI agent is not instantly created; it is not a static concept. Rather, it is a process in which expectations, outcomes, and emotions are involved in altering the trust over time \cite{hoffman2013trust}. The first impression is crucial for the trust trajectory between the human and the AI agent \cite{glikson2020human}. It has been empirically shown that a negative first impression has a more crucial outcome for trust development compared to a situation in which the first impression is positive. However, during the experience, the user sometimes encounters inaccurate outputs which negatively affect the trust \cite{beggiato2013evolution}\cite{nourani2019effects}. Trust formation is essential as trust can break, and one must understand how to recover it in order for users to adopt the AI automation agent. The study of Tolmeijer and colleagues \cite{tolmeijer2021second} addresses the question of how to regain trust with the manipulation of AI-agent accuracy. Even if on first impression the outputs are wrong, it has been found that trust can be recovered if the AI agent will result in the correct output for the user in the near future. Moreover, the importance of first impression has been reflected in the previous study and was first explored in a multi-session experimental setting.  

In the context of LLM-based automation, we highlight the following main considerations:  

\begin{itemize}
     \item \textbf{[TT-SG]: Safety Guardrails}.  The agent includes bulletproof safety guardrails so it can never fail and lose trust.  Examples could include (1) the use of a white list of approved actions by the user in specific situations; (2) the use of customized knowledge created by the end user herself; (3) the ability to customize and teach the agent how to respond to different situations. 
    \item \textbf{[TT-FG]: Fail Gracefully}.  The ability of the agent to always offer mechanisms to restore trust when undesirable situations happen.  For example, if the agent made a mistake, offer a way to not make the same mistake ever again. 
   
\end{itemize}

\section{Assessment}
\label{sec:assessment}
 
In this section, we present an initial assessment of the aforementioned trust characteristics in nascent AI-based automation agent products.  We first provide a detailed concrete example of how these characteristics are evident in a one of these products. Next, we present a comparative analysis that measures the extent to which these concerns are already reflected in the state of the practice products. 

\subsection*{ChatGPT - Example}
ChatGPT is a state-of-the-art conversational artificial intelligence LLM. It enables interactive and natural language conversations by generating coherent responses in various domains. Since ChatGPT was first open to the general population use, its popularity has increased dramatically. We start with an example of our six dimensions with the ChatGPT system. 

\begin{itemize}
    \item 
    \textbf{[Openness]}: When a user first creates an account in ChatGPT, he's presented with several welcome messages that (1) share the goal of the agent \textit{"Our goal is to get external feedback in order to improve our systems and make them safer"}; (2) Provide information about reliability \textit{"While we have safeguards in place, the system may occasionally generate incorrect or misleading information and produce offensive or biased content. It is not intended to give advice."}.  
    \item
    \textbf{[Reliability]}: When a user creates a new chat, ChatGPT discloses: \textit{ChatGPT may produce inaccurate information about people, places, or facts}.
    \item
    \textbf{[Tangibility]:} ChatGPT has a chat UI, however, it does not include an Avatar.  The only visual representation is the model which is being selected - either GPT3.5 or GPT4.  
    \item 
    \textbf{[Immediacy Behavior]:} If the user tells ChatGPT that it has made a mistake, it will respond with empathy \textit{Apologies for the misunderstanding}.  
    \item 
    \textbf{[Task characteristics]:} ChatGPT plugins offer 3rd party vendors to provide APIs to external knowledge and systems.   The APIs\cite{ChatGPTDocs} and safety documentation\cite{ChatGPTSafety} provide guidelines to developers on how to write the plugins.   Some mechanisms for safety have been designed to support human-in-the-loop content moderation and guidelines for open-ended tasks.  In addition, OpenAI performs a review of 3rd party plugins. During the technical review, plugins will receive one of the following states: approved, unverified, banned.
    \item
    \textbf{[Trust trajectory]:}  Some of the messages for building trust are shown for first-time users, or during initial sessions with the agents.  As people become more familiar with the capabilities, they have less need to see these messages again and again.

\end{itemize}

In addition to the standard capabilities, plugins for ChatGPT are now available as a beta feature. They are intended to enable ChatGPT to interact with APIs such that for example, it will be able to assist users with actions such as hotel booking.


\subsection{Comparative Analysis}  

In this section, we provide a comparative summary of how nascent products in the LLM-based automation agent category, address the trust considerations that were highlighted in this paper.  We compare chatGPT+plugins \cite{ChatGPT-Plugins}, MSFT 365 copilot \cite{MicrosoftCopilot}, Adept.AI \cite{AdeptAct1}, and AgentGPT \cite{AgentGPT}.  The goal of the comparison is mainly to understand to which extent recent products are already addressing these concerns.
We note that this is not a detailed comparison and might not reflect the entire abilities of the tools but rather our own opinion after trying these models. On some dimensions, we state only part of the special considerations.

For the purpose of presenting a concise comparison, we have merged the prompt and output mediation, as well as the grounding considerations.  We didn't include the ethics and legal dimension, nor the style adaptation consideration which were difficult to assess.  For each dimension, we mark with either \textbf{[yes]} when we see the product addressing the dimension, \textbf{[no]} when we don't, and \textbf{[-]} when we either don't know or if this is not relevant for the product.  The comparison results are presented in \ref{tbl1} and \ref{tbl2}.  For each consideration, we mark the assessment ranks for the compared products, and detail a concrete example of how this consideration is handled in ChatGPT+plugins system:

\begin{table}
\label{tbl1}
  \footnotesize
  \caption{Trust considerations in reliability and openness}
  \begin{tabular}{lccccccc}
    \toprule
    \multicolumn{1}{c}{} &
    \multicolumn{4}{c}{\textbf{Reliability}} &
    \multicolumn{3}{c}{\textbf{Openness}} \\
    \multicolumn{1}{c}{\textbf{Product}} & \multicolumn{1}{c}{\textbf{Mediation}} & \multicolumn{1}{c}{\textbf{Grounding}} & \multicolumn{1}{c}{\textbf{Feedback}} & \multicolumn{1}{c}{\textbf{Testing}} & \multicolumn{1}{c}{\textbf{Reliability}} & \multicolumn{1}{c}{\textbf{Goal}} & \multicolumn{1}{c}{\textbf{Algorithm}}   \\
    \midrule
    ChatGPT+plugins   & Yes & Yes  & Yes  & No & Yes  & Yes  & Yes \\
    MS Copilot   & Yes & Yes  & -   & - & Yes  & Yes  & Yes \\
    AgentGPT   & No & No  & No  & No & No  & Yes  & Yes \\
    Adept.ai   & - & No  & No  & No & No  & Yes  & - \\
    \bottomrule
  \end{tabular}
\end{table}

\begin{table}
\label{tbl2}
  \footnotesize
  \caption{Trust considerations in tangibility, immediacy behavior, task characteristics, and trust trajectory}
  \begin{tabular}{lccccccc}
    \toprule
    \multicolumn{1}{c}{} &
    \multicolumn{1}{c}{\textbf{Tangibility}} &
    \multicolumn{1}{c}{\textbf{Immediacy behaviors}} &
    \multicolumn{3}{c}{\textbf{Task Characteristics}} &
    \multicolumn{2}{c}{\textbf{Trust trajectory}} \\
    \multicolumn{1}{c}{\textbf{Product}} & 
    \multicolumn{1}{c}{\textbf{T-VR}} & 
    \multicolumn{1}{c}{\textbf{I-Empathy}} & \multicolumn{1}{c}{\textbf{TC-HLM}} & \multicolumn{1}{c}{\textbf{TC-AA}} & \multicolumn{1}{c}{\textbf{TC-OET}} & \multicolumn{1}{c}{\textbf{TT-SG}} & \multicolumn{1}{c}{\textbf{TT-FG}}   \\
    \midrule
    ChatGPT+plugins   & No & Yes  & Yes   & Yes & Yes  & -  & - \\
    MS Copilot   & No & -  & Yes   & - & -  & Yes  & - \\
    AgentGPT  & No & -  & Yes   & No & Yes  & No  & No \\
    Adept.ai   & No & No  & -   & Yes & No  & -  & - \\
    \bottomrule
  \end{tabular}
\end{table}

\textbf{1. Reliability} as detailed in Subsection \ref{Reliability}, refers to the ability of the AI automation agent to be consistent in its output and, therefore, to create the perception of predictability to the user. Our considerations that were assessed under this dimension are (1) \textit{prompt mediation}, (2) \textit{task grounding}, (3) \textit{user feedback}, and (4) \textit{testing}. (1) \textit{prompt mediation} refers to the ability of the AI automation agent to check and validate that the model will answer optimally to the user's intent by optimizing the prompt that serves as the first input from the user. Prompt mediation is conducted in chatGPT+plugins as the prompt is processed and optimized under the hood before it reaches the LLM, calling an appropriate API and eventually producing an action. (2) The \textit{task grounding} aspect refers to the ability of the AI automation agent to validate its abilities compared to the user intention. ChatGPT+plugins has this property, as, for example, if a base of knowledge is missing, the output of the agent will reflect its inability rather than making decisions that can be unrelated to the user's intention. Moreover, the model will not choose APIs on its own but will ask for user approval instead. (3) \textit{User feedback} consideration refers to the adaptive change in the AI automation agent in the future by taking user feedback of past experience into account. ChatGPT has this ability as the model changes based on feedback from the user. The API is revealed and adapted by the user. Finally, (4) \textit{testing} relates to the ability to check before the automation execution of its output and process. In ChatGPT+plugins this is not the case; there are no tools to assess it before the execution.

\textbf{2. Openness}, as detailed in Subsection \ref{openess}, refers to the ability of the AI automation agent to be transparent so the user can understand how the AI works and how much it is accurate and reliable. Our special considerations that were assessed under these dimensions are: (1) \textit{reliability}, (2) \textit{goals}, and (3) \textit{algorithms}. (1) \textit{reliability}, in the context of openness, refers to the ability of the AI automation agent to rate its own reliability. ChatGPT+plugins reflects its ability such that the user can assess it and adopt its use accordingly. There is knowledge of what kind of APIs are available and their intentions. (2) \textit{Goals} refers to whether the vendor reveals his intentions for the product. In chatGPT+plugins this is known, as the product is open access, and the user is familiar with the planner's intention for the product. Finally, (3) \textit{algorithms} assessment refers to the question of how the algorithms work and if they are available and presented to the user. In ChatGPT+plugins, the user receives the information of what data was collected for the model and presents a general explanation of how it works for the AI-naïve user. It is important to note, that though there is information for AI-naïve users, any experts in the field do not have any access to knowledge of its architecture now how specifically the algorithms work.

\textbf{3. Tangibility} as detailed in Subsection \ref{Tangibility}, refers to an AI automation agent's physical or visual/auditorial presence. Our special consideration that was assessed under this dimension is visual presence. ChatGPT has an icon and a visual virtual environment similar to WhatsApp or other familiar human-to-human chat environments, but this is still far from making it look like a person or having gestures like a human.

\textbf{4. Immediacy behaviors} as detailed in Subsection \ref{Immediacy behaviors}, are the perceived physical and/or psychological closeness between people, and in the context of a virtual AI agent, these behaviors serve as the basis for the anthropomorphism of the virtual AI agent. Our special consideration that was assessed under this dimension is empathy, meaning the extent to which the AI automation agent interacts with the user in a way that considers the user's feelings. ChatGPT has this ability as when a mistake happens and the user report on it, ChatGPT will respond with “sorry... / my apology.”

\textbf{5. Task characteristics} as detailed in Subsection \ref{Task Characteristics}, refer to the nature of the task and its sensitivity to the user. Our special considerations that were assessed under this dimension are (1) \textit{human-in-the-loop moderation} (2) \textit{autonomous actions}, and (3) \textit{open-ended tasks}. (1) \textit{human-in-the-loop moderation} is the magnitude in which the user can control and be involved in the AI autonomous agent actions. ChatGPT+plugins uses human-in-the-loop moderation for the choice of API. This is a point at which the user can interfere with the process until he is satisfied. (2) \textit{autonomous actions} are actions that will be done in automation and can include high-sensitivity tasks performed for the user. ChatGPT+plugins can perform autonomous actions and therefore those actions are highly sensitive to user perceptions of accuracy when it comes to trust. (3) \textit{open-ended tasks} are actions with no clearly defined correct output. ChatGPT+plugins include such tasks. To influence trust level in such cases, one can, for example, provide the user with several optional outputs that the system can return, or the system can disclose its output diversity to reflect its future actions in diverse scenarios explicitly. 
 
\section{Discussion}
\label{sec:discussion}

In this paper we argue that AI systems must be developed with trust considerations, originating from human-to-human trustworthy interaction. Reliability, openness, tangibility, immediacy behavior, and task characteristics have been found to be core considerations and are becoming eminent in nascant LLM-based agent systems. These systems need to also adhere to the highest ethical standards and operate in a predictable, understandable manner that aligns with human values. The emergence of LLMs that can understand and produce human-like language introduces a plethora of opportunities but also significant challenges for cultivating and maintaining trust.

We have highlighted the critical role of trust in successful interactions and in the adoption of AI agents, especially in the face of increasingly autonomous AI systems. We have explored trust from both a cognitive and emotional perspective and projected its applicability to human-to-AI-agent interactions. We have also touched upon the key factors influencing trust in AI, discussing their implications for AI in business process automation and automation agents.

One intriguing implication that could arise from this paper is the necessity for all-encompassing metric framework to assess trust in LLM-based Automation Agents.  It is not enough to merely rely on an intuitive understanding of trustworthiness. We need a clear, overall metric framework, akin to a trust maturity model, as well as more visual and interpretable indicators to make trustworthiness more tangible to end users. Within this context, crucial questions revolve around defining the metrics, operationalizing their measurements, and determining the responsible party for certifying these metrics.  We call for new methodologies and tools for the testing and validation of trust in AI agents. Potentially an interdisciplinary approach, engaging researchers from Human-Computer Interaction, User Experience, AI, and related fields, could help tackle these challenges more holistically.

This paper serves as a call to action for these communities. As AI continues to exponentially evolve and defuse into every aspect of our lives, building and maintaining trust in AI is not just a technical challenge but a societal one that will determine the level of adoption of these technologies.  It requires a coordinated, collective effort from all stakeholders involved to ensure that the AI future we're building is one we can all trust.

Future work will also discuss the question of what if humans give a high level of trust, up to the point that no human judgment will be required for certain tasks. In the field of psychology, there is evidence of what can happen when there are extreme levels of trust, in the context of human-to-human interaction. A famous and extreme experimental example of that is the Milgram studies \cite{milgram1963behavioral} in which people trusted an experimenter up to the point in which they totally obeyed him. The obedience of the subject in the experiment allowed them to conduct a harmful act on another human being (electrical shock). Will humans give extreme trust to AI tools such that they will use them for harm without fully thinking it through? As the AI field evolves, how can we prevent that?

\bibliography{main}

\appendix

\end{document}